\documentclass[iicol, sn-apa]{sn-jnl}

\usepackage{graphicx}%
\usepackage{multirow}%
\usepackage{amsmath,amssymb,amsfonts}%
\usepackage{amsthm}%
\usepackage{mathrsfs}%
\usepackage[title]{appendix}%
\usepackage{xcolor}%
\usepackage{textcomp}%
\usepackage{manyfoot}%
\usepackage{caption}
\usepackage{diagbox}
\usepackage[export]{adjustbox}
\usepackage{listings}%
\usepackage{epsfig}
\usepackage{makecell}
\usepackage{float}
\usepackage{booktabs}%
\usepackage{algorithm}%
\usepackage{algorithmicx}%
\usepackage{algpseudocode}%
\newcommand{\blue}[1]{{\color{black}{#1}}} %
\newcommand{\red}[1]{{\color{black}{#1}}} %
\newcommand{\orange}[1]{{\color{black}{#1}}} %

\begin{document}
\title[Article Title]{DiffuVolume: Diffusion Model for Volume based Stereo Matching}


\author[1]{\fnm{Dian} \sur{Zheng}}\email{zhengd35@mail2.sysu.edu.cn}

\author[1]{\fnm{Xiao-Ming} \sur{Wu}}\email{wuxm65@mail2.sysu.edu.cn}

\author[1]{\fnm{Zuhao} \sur{Liu}}\email{liuzh327@mail2.sysu.edu.cn}

\author[1]{\fnm{Jingke} \sur{Meng}}\email{mengjke@gmail.com}

\author[1,2]{\fnm{Wei-Shi} \sur{Zheng}}\email{wszheng@ieee.org}

\affil*[1]{\orgdiv{School of Computer Science and Engineering}, \orgname{Sun Yat-sen
University}, \country{China}}


\affil[2]{\orgdiv{Key Laboratory of MachineIntelligence and Advanced Computing}, \orgname{Ministry of Education}, \country{China}}


\abstract{Stereo matching is a significant part in many computer vision tasks and driving-based applications. Recently cost volume-based methods have achieved great success benefiting from the rich geometry information in paired images. However, the redundancy of cost volume also interferes with the model training and limits the performance. To construct a more precise cost volume, we pioneeringly apply the diffusion model to stereo matching. Our method, termed DiffuVolume, considers the diffusion model as a cost volume filter, which will recurrently remove the redundant information from the cost volume. Two main designs make our method not trivial. Firstly, to make the diffusion model more adaptive to stereo matching, we eschew the traditional manner of directly adding noise into the image but embed the diffusion model into a task-specific module. In this way, we outperform the traditional diffusion stereo matching method by 22$\%$ EPE improvement and 240 times inference acceleration. Secondly, DiffuVolume can be easily embedded into any volume-based stereo matching network with boost performance but slight parameters rise (only 2$\%$). By adding the DiffuVolume into well-performed methods, we outperform all the published methods on Scene Flow, KITTI2012, KITTI2015 benchmarks and zero-shot generalization setting. It is worth mentioning that the proposed model ranks 1$^{st}$ on KITTI 2012 leader board, 2$^{nd}$ on KITTI 2015 leader board since 15, July 2023.}



\maketitle

\begin{figure*}[t]
    \centering
    \includegraphics[width=0.95\linewidth]{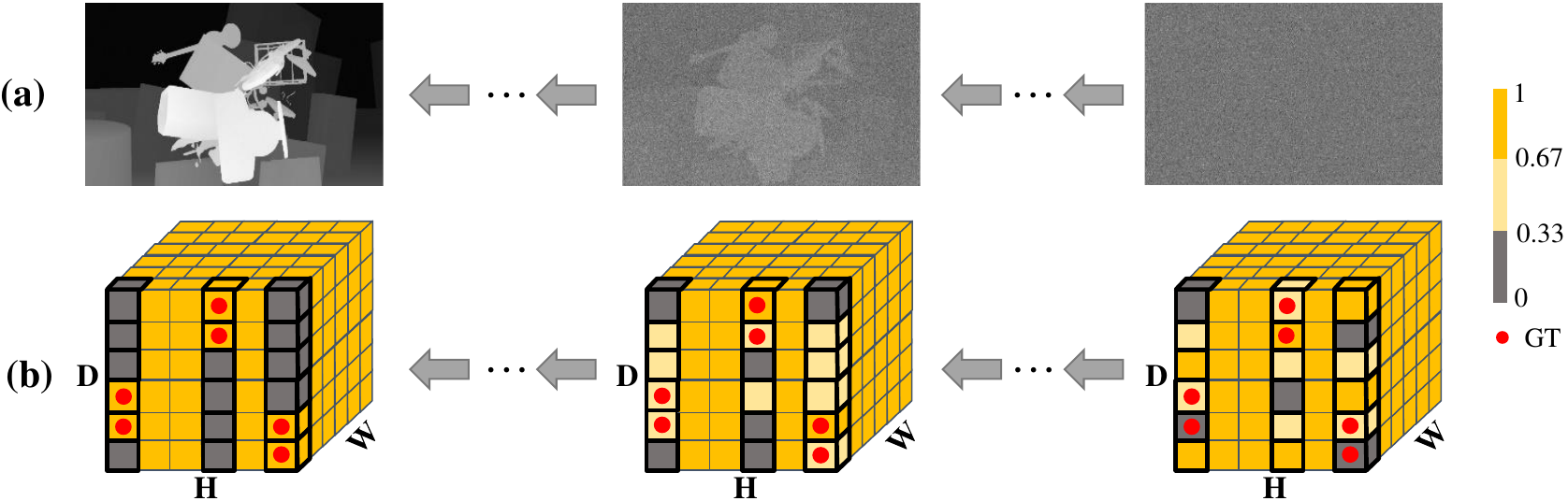}
    \caption{(a) The denoising process of the traditional diffusion model. (b) The redundant information filtering process of our DiffuVolume. The yellow volumes are the filtered volumes at different timesteps, and each column represents the disparity information of one pixel. The red sign means the ground truth disparity levels. The cubes with different colors represent the percentage of reserved information (\textit{i.e.}, yellow, dark gray mean the information in this cube is reserved or filtered entirely). Taking the three \textbf{bold} columns as examples, the redundant information is gradually filtered out as the filtering process goes on, which is similar to the denoising process in diffusion.}
    \label{fig:motivation}
\end{figure*}

\section{Introduction}

\label{sec:intro}
Stereo Matching aims to generate a dense disparity map from a pair of rectified stereo images, which has a wide range of applications in autonomous driving (Chen et al.,~\citeyear{2015deepdriving}) and robotics navigation (Biswas et al.,~\citeyear{2011robot}), \textit{etc.} \par

Recently proposed stereo matching methods (Chang et al.,~\citeyear{2018PSMNet}; Cheng et al.,~\citeyear{cheng2020lea}; Gu et al.,~\citeyear{2020cascade}; Guo et al.,~\citeyear{GWCNet}; Li et al.,~\citeyear{CRE}; Shen et al.,~\citeyear{2021CFNet}; Shen et al.,~\citeyear{pcwnet}; Xu et al.,~\citeyear{2022ACVNet}) mainly use cost volume-based architecture, which aggregates the geometry information from the feature of paired images. Benefiting from the rich geometry information in cost volume, these methods achieve great performance. However, the aggregated information is redundant \red{(Xu et al.,~\citeyear{2022ACVNet})}, since it simply concatenates the whole geometry information in massive feature dimensions. Commonly predicting the disparity of one point does not need such an amount of information, and such redundant information can have a negative impact on interfering with the model training. For example, an image of a wall, which is non-texture (\textit{i.e.}, the adjacent points are highly similar). One point in the left image may find several matched points in the right image, which will confuse the model. In this case, the similar points are redundant information that will ruin the stereo matching training.

In this work, we attempt to remove the redundant information in cost volume by generating an information filter. Intuitively speaking, the ground truth disparity map is a great filter as it contains the correct disparity value and the information in other disparity levels will be removed totally. However, the ground truth disparity map is unavailable \blue{during} the testing stage, so we need to find one with similar effect. Inspired by the nature of the diffusion model, which denoises the Gaussian noise into the desired objective, we observe the consistency that both denoising and filtering aim to remove the redundant information from the target. Based on the observation, we cast the information filtering process as the denoising process as shown in Fig.~\ref{fig:motivation}. The redundant information in cost volume will be recurrently filtered, which is similar to denoising the noise in the image. \red{We validate the redundant information filtering ability and the consistency in Sec~\ref{sec:4.3.1}.}
\par

With the above idea, we design a diffusion model for volume-based stereo matching, termed DiffuVolume, which contains two novel designs. Firstly, we embed the diffusion model into a task-specific module(\textit{e.g.}, cost volume). The consideration is based on two observations: 1$)$ Diffusion model can only recover the data distribution but not the accurate pixel value, which is unsuitable for stereo matching like dense prediction tasks; 2$)$ traditional image-denoising-based diffusion methods \red{(Ho et al.,~\citeyear{ddpm}, Song et al.,~\citeyear{ddim})} require U-Net \red{(ronneberger et al.,~\citeyear{ronneberger2015unet})} with large computing complexity. It is resource-intensive both in the training and inference stages \red{for stereo matching}. Following the observations, we construct an attention-like diffusion filter discretized from the disparity map and multiply it into base cost volume as shown in Fig.~\ref{fig:train}. The obtained filtered volume will predict the disparity map as usual. In this process, we alter the diffusing object from the image to an attention-like diffusion filter. As we discretize each point to a probability vector, which contains more information compared with a pixel, we mitigate the flaw of inaccurate pixel prediction in image-denoising-based diffusion methods.
What's more, in our design, the diffusion filter can be discretized by the disparity map and the disparity map can further transform to the noise. So we do not need the U-Net to predict the noise to renew the diffusion filter, but a lightweight stereo matching network to predict the disparity map only. In this way, we accelerate the inference speed by 240 times compared with the traditional diffusion-based stereo matching method (Shao et al.,~\citeyear{2022diffustereo}). Finally, we analyze the theoretical basis of the optimization objective modification (\textit{e.g.}, noise $\rightarrow$ disparity map), and prove that they are essentially equivalent.

\par

\blue{Secondly,} our diffusion filter is a lightweight plug-and-play refinement module, which can be applied to any volume-based stereo matching network. With slight increase  of parameters (only 2\%), we obtain 11\%, 5.2\%, 18\% and 4.6\% EPE improvement on PSMNet (Chang et al.,~\citeyear{2018PSMNet}), GwcNet (Guo et al.,~\citeyear{GWCNet}), RAFT-Stereo (Lipson et al.,~\citeyear{lipson2021raft}) and ACVNet (Xu et al.,~\citeyear{2022ACVNet}), respectively. It is worth mentioning that we can refine all the existing volume-based models. By embedding our diffusion filter into existing well-performed methods, we achieve state-of-the-art performance on several benchmarks compared with all the published work and the rank 1$^{st}$ on KITTI2012 leader board, 2$^{nd}$ on KITTI 2015 leader board \red{since} 15, July 2023. We further perform zero-shot generalization on real-world scenes and achieve state-of-the-art performance.  
\par

In summary, our main contributions are: \par
$(1)$ An effective diffusion-based stereo matching network DiffuVolume is proposed, which casts the information filtering as the denoising process of the diffusion model. In this way, we recurrently filter out the redundant information in cost volume and achieve state-of-the-art performance on several benchmarks.  \par
$(2)$ We propose a novel application of diffusion model by totally embedding the diffusion model into a task-specific module. We alter the diffusing objective from the pixel into a probability vector, eschew the U-Net network and modify the training objective from the noise into the disparity map. Compared with the traditional diffusion-based stereo matching method, we achieve greater performance while avoiding the extensive training for U-Net and unacceptable inference time (265s $\rightarrow$ 1.1s).
\par
$(3)$ The DiffuVolume we proposed is a plug-and-play method, which can recurrently improve any volume-based stereo matching network while only adding less than 2$\%$ of parameters.  \par

\section{Related Work}
\label{sec:relate}

\subsection{Volume-based Stereo Matching}
\label{sec:2.1}
The general paradigm of stereo matching consists of four steps: feature extraction, cost volume construction, cost aggregation, and disparity regression. Among them, cost volume construction is the most important module. Traditional stereo matching methods (Klaus et al.,~\citeyear{klaus2006segment}; Mei et al.,~\citeyear{mei2013segment} and Hirschmuller et al.,~\citeyear{H2007Stereo}) use the sliding window to compute the matching cost with hand-crafted features and optimize each step individually, causing local optimal problems. Recently CNN-based deep learning methods (Cheng et al.,~\citeyear{cheng2020lea}; Gu et al.,~\citeyear{2020cascade}; Guo et al.,~\citeyear{GWCNet}; Kendall et al.,~\citeyear{GCNet}; Mayer et al.,~\citeyear{DispNet}; Shen et al.,~\citeyear{2021CFNet, pcwnet}; Xu et al.,~\citeyear{2021Bilateral}; Xu et al.,~\citeyear{2022ACVNet}) have achieved great performance by the strong ability of learning-based convolutional neural network and diverse construction way of cost volume. They construct the cost volume by interacting with the information of left and right features $F_l$, $F_r$, which consist of concatenation volume and correlation volume.
\par
DispNet-C (Mayer et al.,~\citeyear{DispNet}) firstly construct the 3D correlation cost volume, which calculates the relationship between paired points but loses much structure and context information. The formula is as follows:
\begin{equation}
\label{eqn:1}
    \textbf{C}_{concat}(d, h, w) = \text{Concat}(\textbf{F}_l(h,w), \textbf{F}_r(h-d,w)).
\end{equation}
RAFT-Stereo (Lipson et al.,~\citeyear{lipson2021raft}) proposes all-pairs correlation volume with an iterative update module, which achieves great zero-shot generalization ability. IGEV (Xu et al.,~\citeyear{xu2023iterative}) adds 4D cost volume into the all-pairs volume which improves the prediction of detailed and edge regions.
\par 
GCNet (Kendall et al.,~\citeyear{GCNet}) proposed 4D concatenation cost volume, which enables the network to learn the relationship between adjacent pixels and adjacent disparity levels from the information-rich but redundant concatenation cost volume. GwcNet (Guo et al.,~\citeyear{GWCNet}) argued that both cost volumes have their own shortcomings and designed the group-correlation cost volume which combines the advantage of the two cost volumes, resulting in better information aggregation. It is the widely used concatenation cost volume, the formula is as follows:
\begin{equation}
\label{eqn:2}
\hspace{-2.4mm}
    \textbf{C}_{corr}(g, d, h, w) = \frac{1}{N_c/N_g}\langle{\textbf{F}^g_l(h,w), \textbf{F}^g_r(h-d,w)}\rangle,
\end{equation}
where $\langle{\cdot}\rangle$ means inner product, $N_c$ is the channel numbers and $N_g$ represents the group numbers, each group calculates correlation within $N_c/N_g$ channels. Cascade (Gu et al.,~\citeyear{2020cascade}) further proposed a multi-scale disparity training method from coarse to fine which manually decreases the searching range of the disparity in each stage. By continuously increasing the resolution and reducing the disparity search range, a high accuracy disparity map is obtained, and the calculation cost is greatly reduced. ACVNet (Xu et al.,~\citeyear{2022ACVNet}) improved the fusion form of two volumes to obtain better information representation ability.

\par
The above work mainly makes special designs to achieve better performance. However, most of them ignore the significance of filtering the \red{redundant} information on cost volume. The ACVNet (Xu et al.,~\citeyear{2022ACVNet}) proposed the attention mechanism to filter the cost volume while requiring three-stage training which is time-consuming. Our DiffuVolume can recurrently filter out the redundant information in the cost volume of any previous work with slight parameters rise.\par

\vspace{2mm}
\subsection{Diffusion Model}
\label{sec:2.2diffusion}
Diffusion model is a new generation model built in the way of Markov Chain, which gradually diffuses the image into standard Gaussian noise in the forward process as follows: 
\begin{equation}
\hspace{10mm}
\left\{
    \begin{array}{lr}
        x_t=\sqrt{\alpha}_tx_{t-1} + \sqrt{1-\alpha_t}\epsilon_{t-1},  &  \\
        \par &\\
        x_t=\sqrt{\overline{\alpha}_t}x_0 + \sqrt{1-\overline{\alpha}_t}\epsilon, &  
    \end{array}
\right.
\label{eqn:forwa}
\end{equation}
where $\alpha$ is a manually designed noise coefficient variation over time $\textit{t}$, $\overline{\alpha}_t=\prod_{i=1}^t\alpha_i$ and $\epsilon$ is the adding noise. \orange{The up formula is the process of one-step diffusing and it could be approximated by the below formula, which could obtain the diffused image of any timesteps by only one step.}
In the reverse process, diffusion model \orange{need to} iteratively denoise from standard Gaussian noise into the clean image as follows:
\begin{equation}
\label{eqn:5}
    p_\theta(x_{t-1}|x_t)=\mathcal{N}(x_{t-1}; \mu_\theta(x_t, t), \sigma_\theta(x_t, t)),
\end{equation}
where $\mu$ and $\sigma$ is the mean and variance of the distribution $x_{t-1}$, $\theta$ is the U-Net (Ronneberger et al.,~\citeyear{ronneberger2015unet}) to predict the noise.
\par
There are two widely used sampling strategies in reverse process: standard sampling strategy DDPM (Ho et al.,~\citeyear{ddpm}) and an acceleration sampling strategy DDIM (Song et al.,~\citeyear{ddim}). For DDPM sampling strategy, it denoises the Gaussian noise step by step, the equations are as follows: 
\begin{equation}
\label{eqn:miu}
    \mu_{\theta}(x_t,t) = \frac{\sqrt{\alpha_t}(1 - \overline{\alpha}_{t-1})}{1 - \overline{\alpha}_t}x_t + \frac{\sqrt{\overline{\alpha}_{t-1}}\beta_t}{1 - \overline{\alpha}_t}x_0^c,
\end{equation}
\begin{equation}
\label{eqn:ddpm}
    x_{t-1} = \sigma_{\theta}(x_t,t)x_t + \mu_{\theta}(x_t,t),
\end{equation}
where $\sigma_{\theta}(x_t,t)$ can be designed manually.   
\par
For DDIM sampling strategy, it proves that adopting skip-step sampling will not influence the theoretical basis of diffusion model. For example, DDIM can sample only 5 steps (\textit{e.g.}, 1000, 800, 600, 400, 200) to simulate the 1000 steps sampling of DDPM. We describe the process in Sec~\ref{sec:reverse}.
\par
As diffusion model can only recover the approximate distribution of the original image. The usage of the diffusion model concentrates on image generation (Saharia et al.,~\citeyear{score_image_generation}; Song et al.,~\citeyear{i2i}), image super resolution (Saharia et al.,~\citeyear{2022IS}), image classification (Yang et al.,~\citeyear{diffusion-classi}), \textit{etc.}, which requires rich semantic information and certain randomness.
\par
\begin{figure*}[t]
    \centering
    \includegraphics[width=1.0\linewidth]{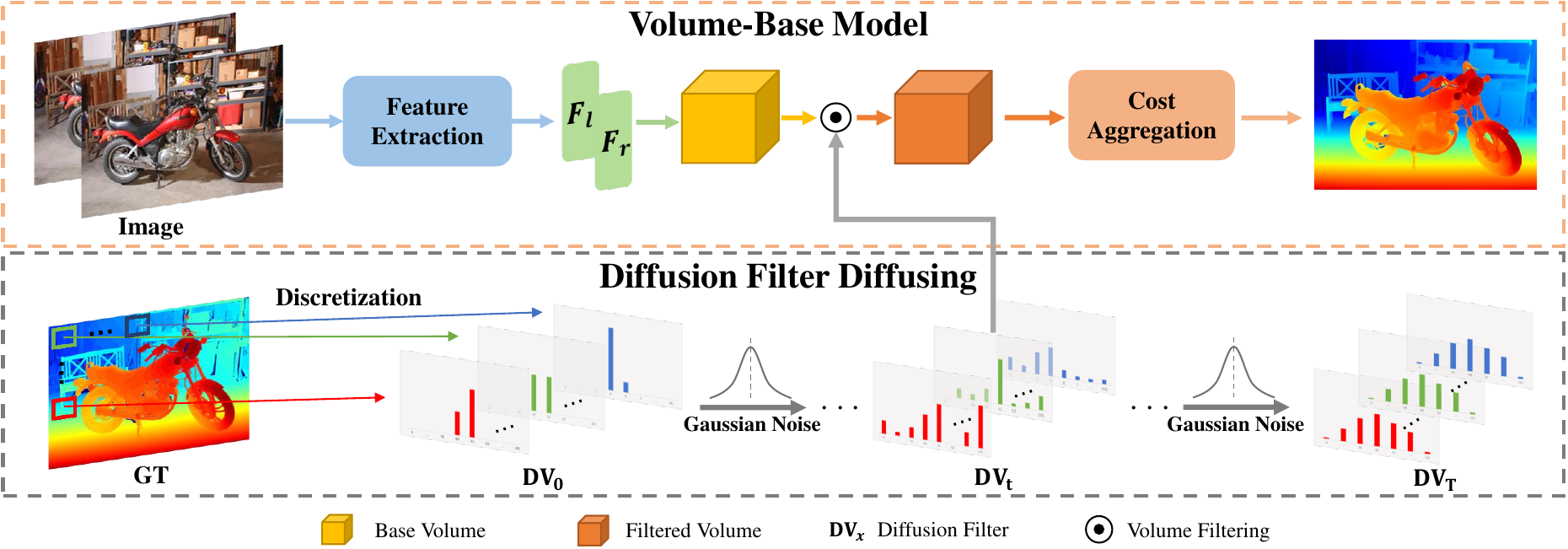}
    \caption{The framework of our proposed DiffuVolume in forward(training) process. The pink one is the base model which can be substituted at will. The gray one is the forward process of our DiffuVolume, each point in the ground truth map will transfer into a probability volume termed DiffuVolume$_0$, $DV_0$ for short. we call the intermediate forward steps $DV_t$ and the final result $DV_T$, which is totally the Gaussian noise. We randomly choose one step and multiply it into the base volume, resulting in the filtered volume.}
    \label{fig:train}
\end{figure*} 

While the diffusion model works well in the field mentioned above, the application on \red{dense} prediction tasks is challenging. These tasks require not only rich semantic information but also accurate pixel value. Some pioneers have tried to introduce the diffusion model into the low-level task. Baranchuk et al.,~\citeyear{baranchuk2021label} applied diffusion into the semantic segmentation by using the pre-trained U-Net model in diffusion way as a strong feature extraction module. DiffuStereo (Shao et al.,~\citeyear{2022diffustereo}) succeeded in stereo matching based 3D human reconstruction which diffuses the residual map, which is the difference between ground truth disparity and coarse disparity map obtained by previous stereo matching methods. The method iteratively refines the coarse disparity map to the accurate disparity map, achieving good performance on the human dataset.
\par

However, DiffuStereo (Shao et al.,~\citeyear{2022diffustereo}) performs poorly on the mainstream of stereo matching. On the one hand, the main dataset of the stereo matching is scene-based which is more complicated than a human (\textit{i.e.}, judging more complex boundaries and distinguishing between foreground and background). On the other hand, to constrain the reverse process of diffusion model, they add extensive conditions to the U-Net network which is complicated. Different from DiffuStereo, the proposed DiffuVolume does not directly diffuse the image, but the attention-like diffusion filter. It recurrently filters out the redundant information in cost volume while keeping the model concise. The extensive experiments show that our DiffuVolume obtains better performance and faster inference speed.


\section{DiffuVolume for stereo matching}
\label{sec:method}
In this section, we describe the detailed designs of integrating the diffusion process into a task-specific module in stereo matching (\textit{i.e.}, cost volume) \blue{to solve the problem of existing redundant information in cost volume and refine the disparity map iteratively}. Then we show the whole network architecture to present that our DiffuVolume is a plug-and-play module with slight parameters rise.

\begin{figure*}[t]
    \centering
    \includegraphics[width=0.95\linewidth]{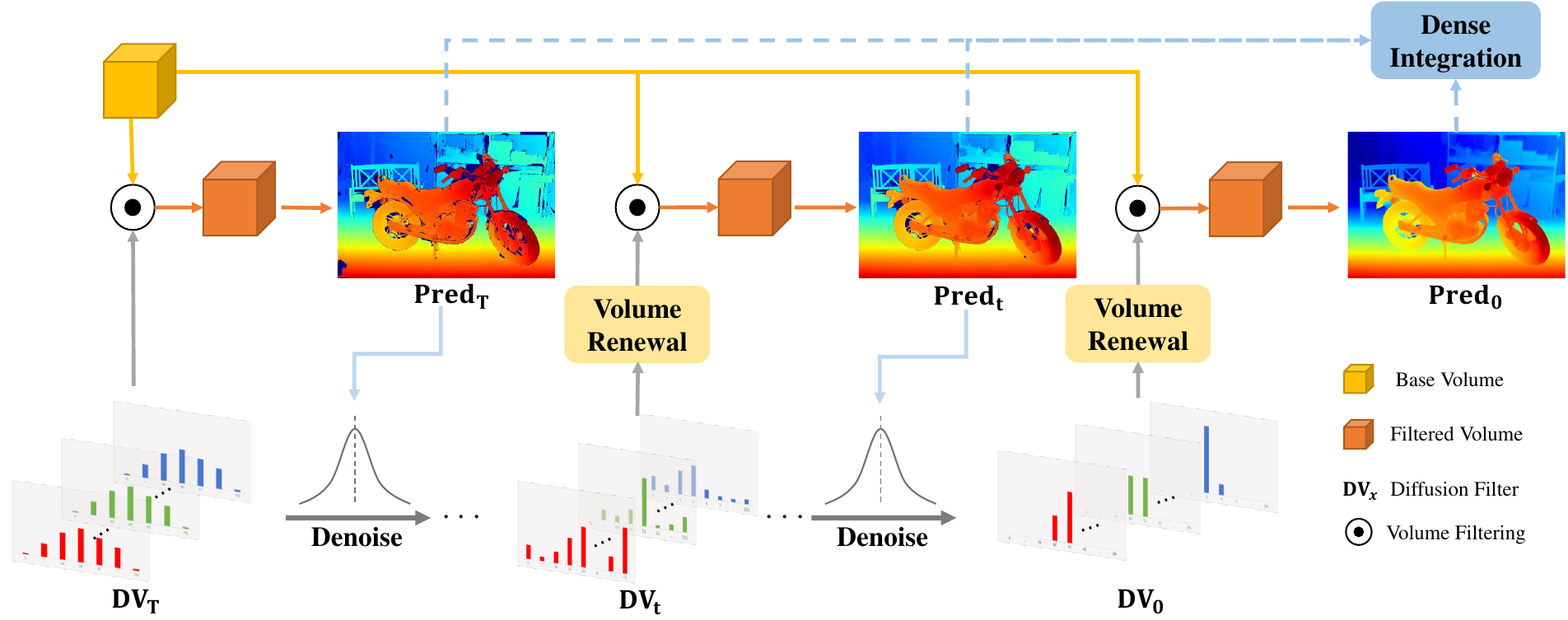}
    \caption{The reverse \red{(inference)} process of our DiffuVolume. The $DV_T$ is set as standard Gaussian noise. \red{It \blue{performs} volume filtering (\textit{i.e.}, element-wise product) with the base volume to obtain the filtered volume. The filtered volume will be fed into the cost aggregation module and obtain the disparity map $Pred_{T}$}. $DV_{T-1}$ can be calculated by the DDIM sampling strategy. \red{$DV_{T-1}$ will perform volume renewal and iteratively filter out the redundant information in base volume to obtain the final prediction $Pred_{0}$}.}
    \label{fig:infer}
    \vspace{-2mm}
\end{figure*}

\subsection{Non-image-diffusing forward process}
\label{sec:train}
In the forward process, we convert the diffusing objective from the image into the attention-like diffusion filter, which contains three steps: filter initialization, filter diffusing and volume filtering. The architecture is shown in Fig.~\ref{fig:train}.
\par

\vspace{0.1cm}

\noindent\textbf{- Filter Initialization.} We first downsample the ground truth disparity map with the size $ H \times W$ into $ H/4 \times W/4$ and discretize each pixel into two integers closest to it with the size $D/4 \times H/4 \times W/4$, which is the initialization of diffusion filter, termed $DV_0$. The process can be computed as:
\begin{equation}
\label{eqn:two-hot}
    DV_{0}(D/4, x, y) = \textbf{Discretize}(d_{gt}(x, y)), 
\end{equation}

\noindent where $d_{gt}$ is the ground truth disparity value, D is the max disparity value which is 192 in training. We take an example to illustrate the operation of discretization. If the disparity value of one point is 30.5, the probability of $30^{th}$ and $31^{th}$ disparity level is set to 0.5 while the rest is set to 0, so the weighted summation of the ${DV}_{0}$ \orange{level} is still 30.5.

\par
\vspace{0.1cm}

\noindent\textbf{- Volume Diffusing.} We alter the diffusing object from the image into the attention-like diffusion filter. To fit the input requirement mentioned in DDPM (Ho et al.,~\citeyear{ddpm}), we rescale the value of the diffusion filter ${DV}_{0}$ from $(0,1)$ into $(-1,1)$ \red{(In the following the signs of $DV_0$ and $DV_0^c$ are after rescaling without extra illustration for sign pithy)}. It is the start of the forward process, then we diffuse it into the Gaussian noise by:
\begin{equation}
\label{eqn:forward}
    DV_t=\sqrt{\overline{\alpha}_t}DV_0 + \sqrt{1-\overline{\alpha}_t}\epsilon,
\end{equation}
where $\alpha_t$ is the noise coefficient, $\overline{\alpha}_t=\prod_{i=1}^t\alpha_i$, $\epsilon$ represents the adding noise and $\textit{t}$ is the timestep.  \par
By changing the diffusing object, we mitigate the inaccurate prediction of diffusion model.
\par

\vspace{0.1cm}

\noindent\textbf{- Volume Filtering.} In this \red{step}, we filter out the redundant information by diffusion filter. We randomly sample one step t and the $DV_t$ with the size of $1 \times D/4 \times H/4 \times W/4$ will add the time embedding with the size of $1 \times D/4$ to sense different timesteps. Then we do element-wise multiplication with the base cost volume with the size of $C \times D/4 \times H/4 \times W/4$ in channel dimension C, the formula is as follows:
\begin{equation}
\label{eqn:embedding}
    \textbf{C}_{flt} = \textbf{C}_{base} \odot (DV_t + \textbf{MLP}(t)),
\end{equation}
where $\odot$ is the element-wise product, $\textbf{C}_{flt}$ means the filtered volume, $\textbf{C}_{base}$ means the base volume, t is the selected timestep and $\textbf{MLP}$ is the fully connected layer for capturing the information of time sequence. \par


\subsection{Loss function}
\label{sec:loss}
As \blue{our} diffusion filter can be obtained by discretizing the disparity map and the disparity map can further calculate the noise (\textit{i.e.}, described in Sec~\ref{sec:reverse} and Sec~\ref{sec:theory}), we do not need an extensive U-Net network to predict the noise to renew the diffusion filter, but a stereo matching network to predict the disparity map. So we could only use the traditional stereo matching loss function, which calculates the L1 distance between the ground truth $d_{gt}$ and the predicted disparity map $d_{pred}$, the formulation can be written as:
\begin{equation}
     \mathcal{L} = \frac{1}{HW}\sum_{i=0}^2\sum_{x=1}^H\sum_{y=1}^W\lambda_i\Vert d_{gt}(x, y) - d_{pred}^i(x, y) \Vert_1,
\end{equation}
where $\textit{i}$ means the index of the three hourglass outputs in cost aggregation modules and $\lambda$ means the coefficient of each output. 
\par

\subsection{Stereo matching specific reverse process}
\label{sec:reverse}
In the reverse process, DiffuVolume is a denoising sampling process from standard Gaussian noise to desired diffusion filter, which filters redundant information gradually. It consists of three modules: variant of DDIM sampling, volume renewal and \blue{dense integration}, which is shown in Fig.~\ref{fig:infer}. 

\par
\label{sec:infer}

\vspace{0.1cm}

\noindent\textbf{- Variant of DDIM Sampling.} We set standard Gaussian noise as the $DV_T$. $DV_T$ will first multiply into the base volume followed Eq. (\ref{eqn:embedding}) and predict the disparity map $Pred_{T}$. The $Pred_{T}$ will be transformed into probability volume using discretization which can be seen as \textbf{the coarse proximity} to $DV_0$, we name it $DV_0^c$. Then we calculate the added noise in the forward process via the following formula:
\begin{equation}
\label{eqn:coarse0}
    DV_0^c(D/4, x, y) = \textbf{Discretize}(Pred_{T}(x, y)), 
\end{equation}
\begin{equation}
\label{eqn:pred_noise}
    \epsilon= \frac{1}{\sqrt{1-\overline{\alpha}_T}}(DV_T - \sqrt{\overline{\alpha}_T}DV_0^c).
\end{equation}
The formula Eq. (\ref{eqn:pred_noise}) is the variant of Eq. (\ref{eqn:forward}) which is used in DDPM (Ho et al.,~\citeyear{ddpm}) and DDIM (Song et al.,~\citeyear{ddim}). Obtaining the noise $\epsilon$ and the coarse $DV_0^c$, we further use the DDIM sampling algorithm to recover the $DV_{T-1}$, which is the \textbf{Denoise} process in Fig.~\ref{fig:infer}. The formula is:
\begin{equation}
\label{eqn:ddim}
    DV_{T-1} = \sqrt{\overline{\alpha}_{T-1}}DV_0^c + \sqrt{1 - \overline{\alpha}_{T-1} - \sigma^2}\epsilon + \sigma\epsilon^*,
\end{equation}
where $\sigma = \eta \sqrt{(1-\frac{\overline{\alpha}_T}{\overline{\alpha}_{T-1}}) \cdot \frac{1-\overline{\alpha}_{T-1}}{1-\overline{\alpha}_T}}$, $\eta$ is the DDIM sampling coefficient and $\epsilon^*$ is the Gaussian noise. \par
The $DV_{T-1}$ will filter the base volume and follow the process above. The diffusion filter will iteratively approach to $DV_0$ and filter out the redundant information in base volume.
\par

\vspace{0.1cm}

\noindent\textbf{- Volume Renewal.} It is inevitable that there are \blue{poor} inference points during the sampling step, \blue{this is} because on the one hand, the inference acceleration sacrifices the generation quality; on the other hand, the diffusion model only generates the result which is similar to the original data distribution while the accurate value is possibly wrong. To alleviate these problems, we do volume renewal after each DDIM sampling step (\textit{e.g.}, denoise process). Our renewal strategy is based on the assumption that the difference of EPE metric between the recent stereo matching network and the ground truth map is lower than 0.5 pixel. We take ACVNet (Xu et al.,~\citeyear{2022ACVNet}) as an example (\textit{e.g.}, 0.48 EPE), once our result is 1-pixel different from the result of ACVNet at one point, the chances are that is wrong. Based on this assumption, in each sampling step $\textit{t}$, we calculate the absolute difference between the ACVNet and our $Pred_t$ and select the outliers. What's more, following the CFNet (Shen et al.,~\citeyear{2021CFNet}) we introduce the uncertainty into our model to select out the points with undesirable probability distribution. The selected point in the $DV_t$ will be reset to Gaussian noise and rescaled into the range of $(0,1)$.
\par

\vspace{0.1cm}

\noindent\textbf{- \blue{Dense Integration}.} Besides the volume renewal mechanism, we further design the \blue{dense integration} module, which calculates the weighted average of the prediction results \red{in different times} (\textit{e.g.}, $Pred_T$, $Pred_t$ and $Pred_0$) and the result of the previous method (\textit{e.g.}, ACVNet), since the recovering process of the diffusion model can be seen as an easy-to-hard process. Each step may perform extremely well in certain areas while the others act in general performance, so \blue{dense integration} is important. What's more, we select the weights of different steps based on the prior knowledge that the diffusion model iteratively refines the image and the later steps will obtain better results, so we give the later step a higher weight. \red{We validate the assumptions above in Sec~\ref{sec:ablation}.}
\par

\subsection{Network Architecture}
\label{sec:architecture}
The intention of our DiffuVolume is a plug-and-play refinement module for all the volume-based stereo matching networks by filtering the cost volume recurrently in a diffusion way. In the following, we introduce the formal network architecture of a basic cost volume-based model \red{(\textit{e.g.}, Chang et al.,~\citeyear{2018PSMNet}; Guo et al.,~\citeyear{GWCNet} and Xu et al.,~\citeyear{2022ACVNet})} combined with our DiffuVolume.
\par

\vspace{0.1cm}

\noindent\textbf{- Feature Extraction.} The formal methods mentioned above use the shared ResNet-like convolution network as the feature extraction for both left and right images. The convolution outputs are two 320-channel unary feature maps. We name it $\textbf{F}_l$ and $\textbf{F}_r$ with the size of $320 \times D/4 \times H/4 \times W/4$. The downsampling is caused by convolution operation in the feature extraction process. 
\par

\vspace{0.1cm}

\noindent\textbf{- Cost volume Construction.} Based on the two volumes mentioned in Sec~\ref{sec:2.1}, previous work will further fuse the two volumes in different ways, we termed it base volume. Our diffusion filter will further multiply into the base volume in the channel dimension which filters out the disparity level information. The obtained cost volume is called filtered volume.
\par

\vspace{0.1cm}

\noindent\textbf{- Cost volume Aggregation.} The filtered volume will be fed into the cost aggregation module, which consists of several 3D stack-hourglass modules to aggregate the disparity-level information.
\par

\vspace{0.1cm}

\noindent\textbf{- Disparity Regression.} After the aggregation, the filtered cost volume with the size of $C \times D/4 \times H/4 \times W/4$ will directly upsample into $C \times D \times H \times W$ and obtain 1-channel disparity probability volume with the softmax operation. The size of probability volume is $1 \times D \times H \times W$ which means the probability of each point at each disparity level. The probability volume will do weighted summation to predict the disparity with the size $H \times W$.
\par

\begin{figure}[t]
    \vspace{-6mm}
    \centering
    \includegraphics[width=1\linewidth]{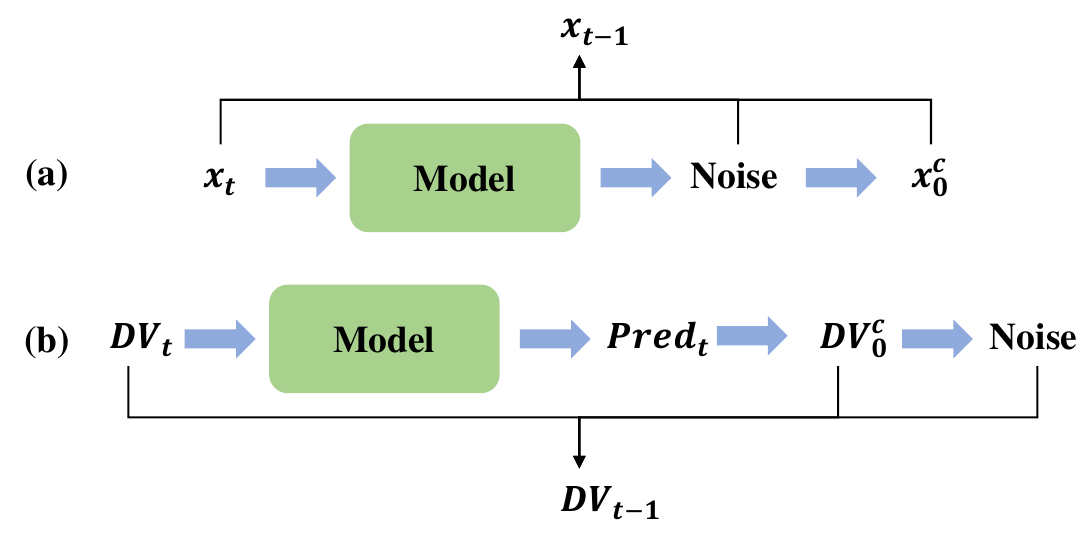}
    \caption{Analysis of the reverse process of (a) traditional diffusion model and (b) ours.}
    \label{fig:ddim_formula}
\end{figure}

\section{Theoretical Analysis}
\label{sec:theory}
In this section, we prove that although we modify the training objective from the noise into the disparity map \blue{to integrate the diffusion process into a task-specific module}, it does not influence the theoretical basis of diffusion. Because it is \textbf{essentially equivalent}, which is also mentioned in the official code of DDPM. We compare our stereo matching specific reverse process with the traditional reverse process in Fig.~\ref{fig:ddim_formula}, the formula of the noise to $x_0$ or $DV_0^c$ to noise is computed as:
\begin{equation}
\label{eqn:8}
    x_0^c = \frac{1}{\sqrt{\overline{\alpha}_t}}(x_t - \sqrt{1-\overline{\alpha}_t}\epsilon),
\end{equation}
\begin{equation}
\hspace{10mm}
\left\{
    \begin{array}{lr}
        DV_0^c(D/4, x, y) = \textbf{Dicretize}(Pred_{t}(x, y)),  &  \\
        \par &\\
        \epsilon= \frac{1}{\sqrt{1-\overline{\alpha}_t}}(DV_t - \sqrt{\overline{\alpha}_t}DV_0^c), &  
    \end{array}
\right.
\label{eqn:chong}
\end{equation}
\par
In Fig.~\ref{fig:ddim_formula}, traditional diffusion methods (\textit{e.g.}, DDPM, DDIM) use the U-Net to predict the noise $\epsilon$ and calculate the coarse $x_0^c$ by Eq. (\ref{eqn:8}), then the $x_0^c$, $x_t$ and the $\epsilon$ is used to recover the $x_{t-1}$, which is similar with the Eq. (\ref{eqn:ddim}).  \par
Our DiffuVolume predicts the disparity map $Pred_{t}$ \red{by the stereo matching network}. The coarse $DV_0^c$ and $\epsilon$ is obtained by Eq. (\ref{eqn:chong}). Finally, the $\epsilon$, $DV_t$ and $DV_0^c$ will be used to recover the $DV_{t-1}$ following Eq. (\ref{eqn:ddim}). \par
The $x_0^c$ in the traditional diffusion formula is the same as $DV_0^c$ in our diffusion network and the $\epsilon$ is the same. Although we modify the training objective from noise $\epsilon$ into disparity map $Pred_{t}$, it does not influence the sampling algorithm in Eq. (\ref{eqn:ddim}) as the main purpose of both networks is predicting the precise $x_0$ by Eq. (\ref{eqn:ddim}). Using the model to predict the noise $\epsilon$ or the $x_0^c$ is essentially equivalent. The theoretical analysis ensures the reliability of our stereo matching specific reverse process.
\par

\begin{table*}[t]
    \centering
    \caption{Quantitative results on Scene FLow dataset. Bold: Best.}
    \setlength{\tabcolsep}{2.5mm}{
        \begin{tabular}{c|ccccccc}
        \hline
        Model & PSMNet & GwcNet & Cascade & CFNet & ACVNet & IGEV
        & DiffuVolume(ours)\\
        \hline
        EPE(\%) & 1.09 & 0.77 & 0.65 & 0.97 & 0.48 & 0.47 & \textbf{0.46} \\
        \hline
        \end{tabular}
        }
    \centering
    \label{tab:scene}
\end{table*}

\begin{table*}[t]
    \centering
    \caption{Quantitative results on KITTI dataset. We report the percentage of the points with errors larger than d pixels in non-occlusion (D-noc) and all(D-all) areas and the EPE both in non-occluded (EPE-noc) and all (EPE-all) areas for KITTI 2012; for KITTI 2015 we report the D1-metric in background (D1-bg), foreground (D1-fg) and all (D1-all) regions. All of the metrics are the lower the better. Bold: Best.}
    \resizebox{2.1\columnwidth}{!}{
        \begin{tabular}{cc|c|cccccc|ccc|c}
        \hline
        \multicolumn{2}{c|}{\multirow{2}{*}{Model}} & \multirow{2}{*}{Year} &
        \multicolumn{6}{c|}{KITTI 2012} & \multicolumn{3}{c|}{KITTI 2015} & \multirow{2}{*}{Run-time(s) $\downarrow$}\\
        \multicolumn{2}{c|}{} & & 2-noc $\downarrow$ & 2-all $\downarrow$ & 5-noc $\downarrow$ & 5-all $\downarrow$ & EPE(noc) $\downarrow$ & EPE(all) $\downarrow$ & D1-bg $\downarrow$ & D1-fg $\downarrow$ & D1-all $\downarrow$ & {} \\
        \hline
        \multicolumn{2}{c|}{SGM} & 2017 & 3.60 & 5.15 & 1.60 & 2.36 & 0.7 & 0.9 & 2.66 & 8.64 & 3.66 & 67\\
        \multicolumn{2}{c|}{GCNet} & 2017 & 2.71 & 3.46 & 1.12 & 1.46 & 0.6 & 0.7 & 2.21 & 6.16 & 2.87 & 0.9\\
        \multicolumn{2}{c|}{PSMNet} & 2018 & 2.44 & 3.01 & 0.90 & 1.15 & 0.5 & 0.6 & 1.86 & 4.62 & 2.32 & 0.41\\
        \multicolumn{2}{c|}{GwcNet} & 2019 & 2.16 & 2.71 & 0.80 & 1.03 & 0.5 & 0.5 & 1.74 & 3.93 & 2.11 & 0.32\\
        \multicolumn{2}{c|}{HITNet} & 2021 & 2.00 & 2.65 & 0.96 & 1.29 & \textbf{0.4} & 0.5 & 1.74 & 3.20 & 1.98 & 0.02\\
        \multicolumn{2}{c|}{CFNet} & 2021 & 1.90 & 2.43 & 0.74 & 0.94 & \textbf{0.4} & 0.5 & 1.54 & 3.56 & 1.88 & 0.18\\
        \multicolumn{2}{c|}{RAFT-Stereo} & 2021 & 1.92 & 2.42 & 0.86 & 1.11 & \textbf{0.4} & 0.5 & 1.75 & 2.89 & 1.96 & 0.38\\
        \multicolumn{2}{c|}{LEAStereo} & 2020 & 1.90 & 2.39 & 0.67 & 0.88 & 0.5 & 0.5 & 1.40 & 2.91 & 1.65 & 0.3\\
        \multicolumn{2}{c|}{ACVNet} & 2022 & 1.83 & 2.35 & 0.71 & 0.91 & \textbf{0.4} & 0.5 & 1.37 & 3.07 & 1.65 & 0.2\\
        \multicolumn{2}{c|}{CREStereo} & 2022 & 1.72 & 2.18 & 0.76 & 0.95 & \textbf{0.4} & 0.5 & 1.45 & 2.86 & 1.69 & 0.4\\
        \multicolumn{2}{c|}{PCWNet} & 2022 & 1.69 & 2.18 & \textbf{0.63} & \textbf{0.81} & \textbf{0.4} & 0.5 & 1.37 & 3.16 & 1.67 & 0.44\\
        \multicolumn{2}{c|}{IGEV} & 2023 & 1.71 & \textbf{2.17} & 0.73 & 0.94 & \textbf{0.4} & \textbf{0.4} & 1.38 & 2.67 & 1.59 & \textbf{0.18}\\
        \multicolumn{2}{c|}{DiffuVolume(ours)} & - & \textbf{1.68} & \textbf{2.17} & \textbf{0.63} & \textbf{0.81} & \textbf{0.4}  & \textbf{0.4} & \textbf{1.35} & \textbf{2.51} & \textbf{1.54} & 0.36\\
        \hline
    \end{tabular}{}
    }
    \label{tab:kitti}
\end{table*}

\section{Experiment}
\label{sec:experiment}
In this section, we first evaluate and visualize the performance of our proposed model on several datasets \textit{i.e.}, Scene Flow (Mayer et al.,~\citeyear{2016Scene}), KITTI2012 (Geiger et al.,~\citeyear{2012KITTI}), KITTI2015 (Menze et al.,~\citeyear{2015KITTI}). We further evaluate our DiffuVolume from four perspectives, validating the ability to filter redundant information in cost volume, evaluating the university as a plug-and-play module, ablating the effectiveness of our DiffuVolume under different designs, and analyzing the failing case of our method. Additionally, we show that our DiffuVolume is more suitable for stereo matching compared with traditional diffusion model. Finally, we do zero-shot generalization on four real-world benchmarks, \textit{i.e.}, KITTI2012 (Geiger et al.,~\citeyear{2012KITTI}), KITTI2015 (Menze et al.,~\citeyear{2015KITTI}), Middlebury (Scharstein et al.,~\citeyear{scharstein2014middu}) and ETH3D (Schops et al.,~\citeyear{schops2017eth3d}).
\subsection{Datasets and Evaluation Metrics}
\noindent\textbf{Scene Flow} is a large synthetic stereo dataset that consists of Driving, Flyingthings3D and Monkaa subsets. The dataset collects 35454 training pairs and 4370 testing pairs with the resolution of 960$\times$540, which provides dense disparity maps as ground truth $d_{gt}$. Following the previous evaluation metrics, we utilized the end-point error (EPE) and the percentage of pixels with errors larger than 1px (bad 1.0). \par
\vspace{2mm}
\noindent\textbf{KITTI} is collected from real-world driving scenes. KITTI 2012 contains 194 training stereo image pairs and 195 testing image pairs, and KITTI 2015 contains 200 training stereo image pairs and 200 testing image pairs. Both datasets provide sparse ground-truth disparity. As the KITTI2012 and KITTI2015 are not big datasets, we fine-tune the model after pre-trained on the Scene Flow dataset. For KITTI2012 we use the evaluation metrics end-point error (EPE) and the disparity outliers percentage D1 and for KITTI2015, the percentage of disparity outliers D1 is evaluated for background, foreground, and all pixels. 
\par
\vspace{2mm}
\noindent\textbf{Middlebury} is a large real-world dataset with 23 paired images under different lighting environments. The largest disparity value can approach 1000. \textbf{ETH3D} is a collection of gray-scale stereo pairs from indoor and outdoor scenes, which contains 27 image pairs. The disparity range is 0-64. As both datasets are too small to train, we perform zero-shot generalization on them. The evaluation metric is the percentage of pixels with errors larger than 2px and 1px respectively. 
\par

\begin{figure*}[t]
    \centering
    \includegraphics[width=0.95\linewidth]{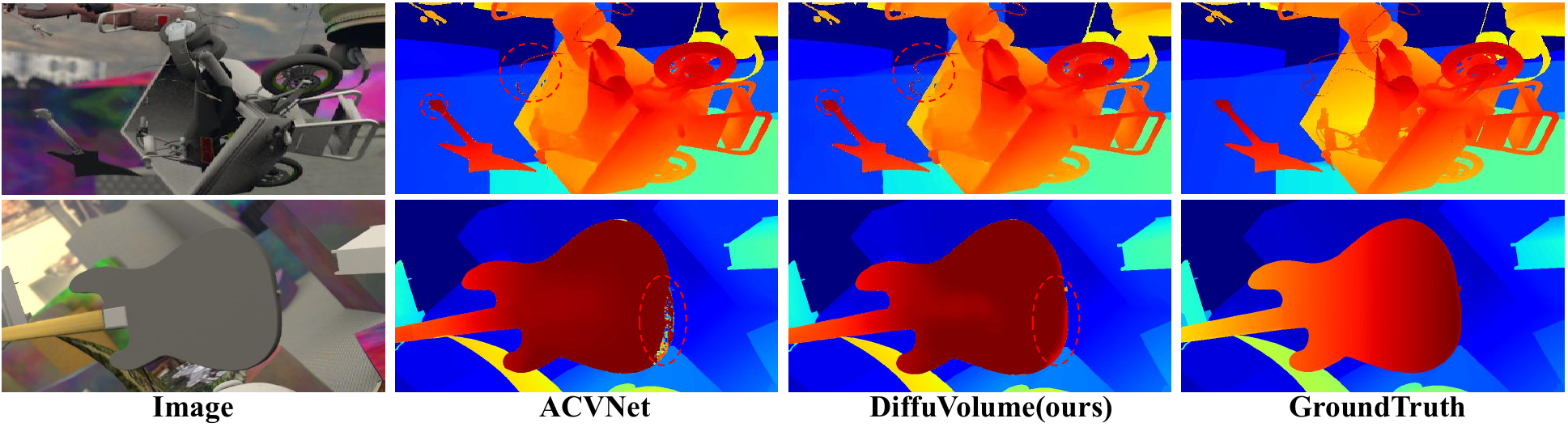}
    \caption{The visualization of our DiffuVolume compared with ACVNet.}
    \vspace{-7mm}
    \label{fig:SceneFlow}
\end{figure*}

\begin{figure*}[t]
    \centering
    \includegraphics[width=0.95\linewidth]{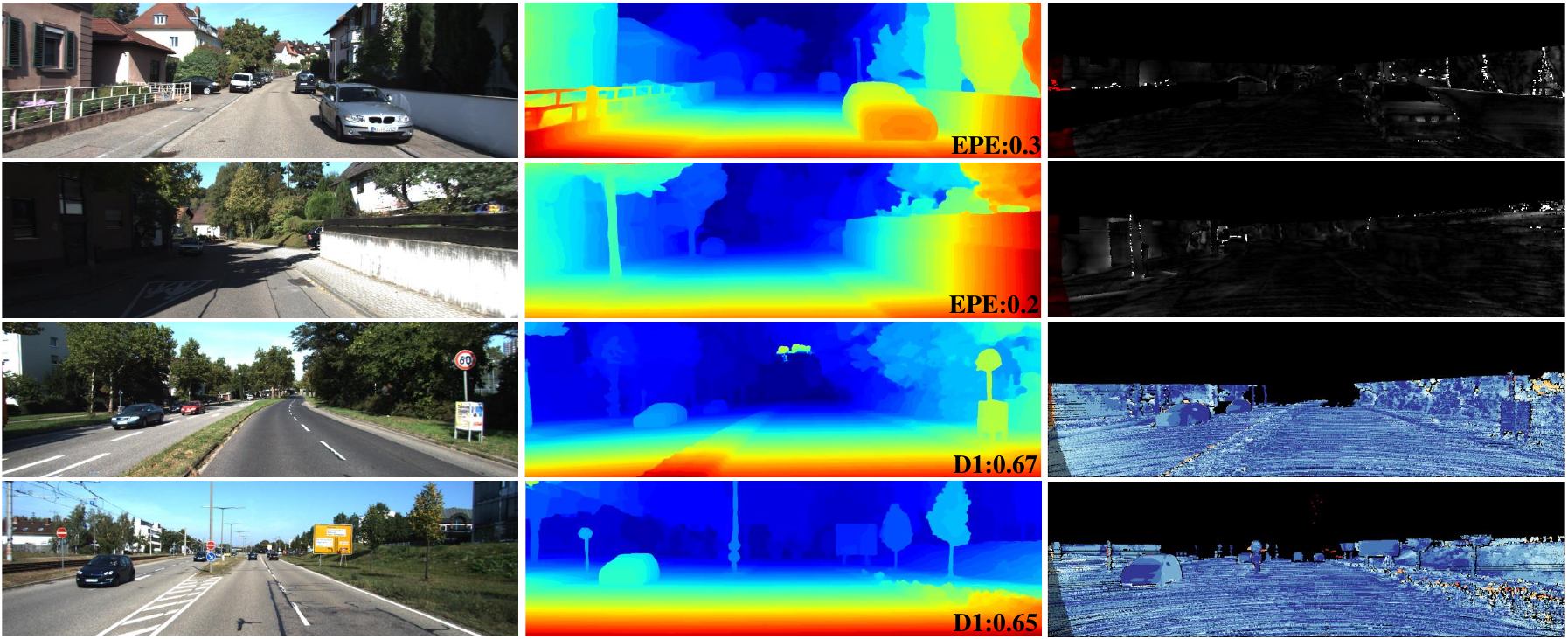}
    \caption{The visualization of our DiffuVolume on KITTI, by column is the RGB, disparity map and the error map respectively. The first two rows are the results of KITTI2012 and the last two rows are of KITTI2015. In the error map, for 2012, the whiter the worse, and the red region is the total occlusion region; for 2015, the bluer the better.}
    \vspace{-4mm}
    \label{fig:kitti}
\end{figure*}

\subsection{Implementation Details}
DiffuVolume is trained using four RTX 3090 GPUs with PyTorch (Paszke et al.,~\citeyear{paszke2019pytorch}) environment. We resize all the images into resolution 512 $\times$ 960. During training, We use the Adam (Kingma et al.,~\citeyear{kingma2014adam}) optimizer and implement our method based on the pre-trained model of the previous cost volume-based network (Chang et al.,~\citeyear{2018PSMNet}; Guo et al.,~\citeyear{GWCNet}; Xu et al.,~\citeyear{2022ACVNet}). We set the timestep as 1000 \orange{and use the DDIM (Song et al.,~\citeyear{ddim}) sampling strategy in the reverse process}. We use the cosine schedule to set noise coefficient $\beta_t, \alpha_t$. We evaluate our model based on different architectures up to different benchmarks, \textit{i.e.}, ACVNet (Xu et al.,~\citeyear{2022ACVNet}) for Scene Flow dataset; PCWNet (Shen et al.,~\citeyear{pcwnet}) for KITTI2012; IGEV-Stereo (Xu et al.,~\citeyear{xu2023iterative}) for KITTI2015 and RAFT-Stereo (Lipson et al.,~\citeyear{lipson2021raft}) for Middlebury and ETH3D. The training schedule follows the original strategy. 

\subsection{Evaluation on the benchmarks}
\blue{In this section, we compare with the recent great works SGM (Seki et al.,~\citeyear{seki2017sgm}); GCNet (Kendall et al.,~\citeyear{GCNet}); PSMNet (Chang et al.,~\citeyear{2018PSMNet}); GwcNet (Guo et al.,~\citeyear{GWCNet}); HITNet (Tankovich et al.,~\citeyear{2020HITNet}); Cascade (Gu et al.,~\citeyear{2020cascade}); CFNet (Shen et al.,~\citeyear{2021CFNet}); RAFT (Lipson et al.,~\citeyear{lipson2021raft}); LEAStereo (Cheng et al.,~\citeyear{cheng2020lea}); ACVNet (Xu et al.,~\citeyear{2022ACVNet}); CREStereo (Li et al.,~\citeyear{CRE}); PCWNet (Shen et al.,~\citeyear{pcwnet}); IGEV (Xu et al.,~\citeyear{xu2023iterative}).}
\par
\noindent\textbf{Scene Flow.} The experiments in Table~\ref{tab:scene} show that we achieve state-of-the-art performance on Scene Flow dataset. We further visualize the results of our DiffuVolume compared with the excellent ACVNet in Fig.~\ref{fig:SceneFlow}. The prediction quality of DiffuVolume in the edge region outperforms the ACVNet and it is worth mentioning that on specific edge regions, DiffuVolume nearly attaches the accuracy of the ground truth.  \par
The results not only ensure that filtering redundant information in volume benefits the model training in the hard areas such as edge and non-texture regions but also validate the effectiveness of our diffusion model. Our result in the first column of Fig.~\ref{fig:SceneFlow} nearly attaches the accuracy of the ground truth map which is impossible for previous stereo matching methods, as this detailed region is hard to predict. As the diffusion process has the potential of perfectly recovering the ground truth, the accuracy in the hard region is high, showing the necessity of filtering the redundant information.
\par

\begin{figure*}[t]
    \centering
    \includegraphics[width=1\linewidth]{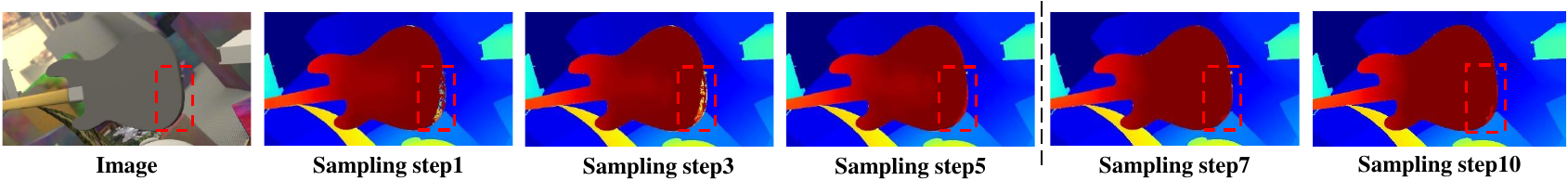}
    \caption{The visualization of one image at different sampling steps. \blue{On the left of the dashed line, as the step grows, the edge region is gradually refined but on the right, over-refining phenomenon occurs}.}
    \label{fig:step}
\end{figure*}

\begin{figure}[ht]
    \centering
    \includegraphics[width=1\linewidth]{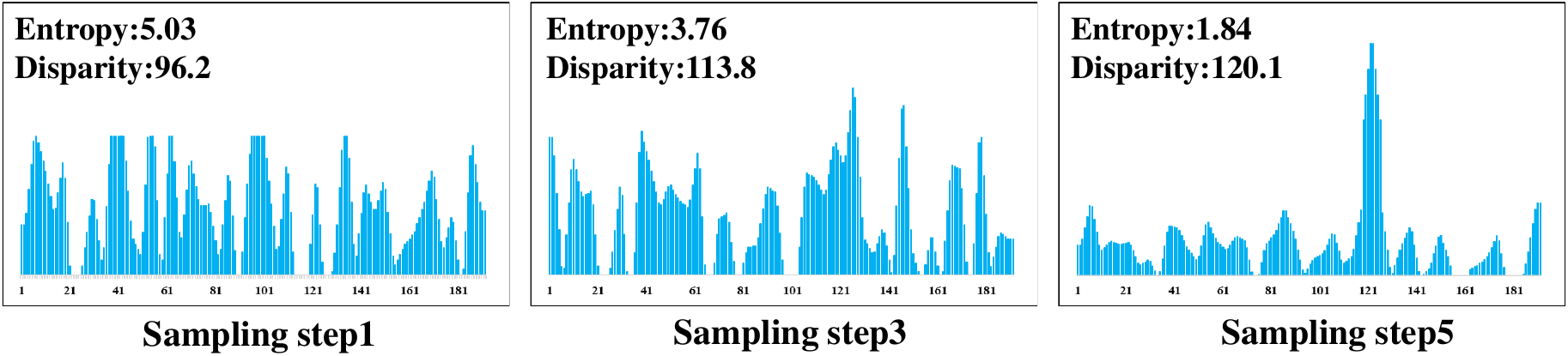}
    \caption{The histogram and the corresponding information entropy of one point in different sampling steps. Each histogram contains the probability at each disparity level. The total step is 5.}
    \label{fig:entropy}
\end{figure}

\vspace{2mm}
\noindent\textbf{KITTI.} As shown in Table~\ref{tab:kitti}, our method outperforms all of the proposed methods both on KITTI2012 and KITTI2015, especially we rank 1$^{st}$ in 5-pixel leader board on KITTI2012 and 2$^{nd}$ on KITTI2015 leader board. We further visualize the results on KITTI2012 and KITTI2015 in Fig.~\ref{fig:kitti}, which is directly obtained from the leader board. The results show the great prediction accuracy of the DiffuVolume.
\par

\subsection{Evaluation of Our DiffuVolume}
\subsubsection{Validating the ability of DiffuVolume to filter the redundant information}
\label{sec:4.3.1}
We introduce information entropy as the evaluation metric to measure the information volume. The formula is:$ H(x) = -\sum_ip(x_i)\log p(x_i)$, where x means one point in the cost volume and i means the disparity level range from 0 to 192. We present the diffusion filter at one point in three different sampling steps as shown in Fig.~\ref{fig:entropy}. The results show that the information entropy is reducing as the iteration grows and the probability vector tends to be a unimodal distribution. The ground truth of this point is 120.3 which fits the probability vector in the last step. The results prove the great ability of DiffuVolume to filter the redundant information in cost volume while maintaining the useful part. The results are the same as our motivation which is shown in Fig.~\ref{fig:motivation}, validating the existence of information redundant phenomenon and the reliability of our novel application of the diffusion process.  \par
We also show the visualization results of our DiffuVolume in different sampling steps in Fig.~\ref{fig:step}. Within 5 steps, as the step grows, the diffusion filter gradually removes the redundant information in cost volume, recurrently refining the edge region in the \textcolor{red}{red} box. The result is consistent with the result of information entropy. However, we observe that for more sampling steps, the model \textbf{over-refines} the edge region. 7$^{th}$ and 10$^{th}$ steps over-refine the edge of the guitar based on the right prediction of 5$^{th}$ step, which inversely ruins the performance. With the interesting observation, we assume that when using the DDIM sampling strategy, the step number of diffusion model in \textbf{dense prediction task} is not the larger the better, the more suitable the better.

\subsubsection{University of DiffuVolume}
The DiffuVolume we propose is a plug-and-play module for all the cost volume-based stereo matching networks. We show the university of our model in Table~\ref{tab:plug} by refining on four state-of-the-art models, \textit{i.e.}, PSMNet (Chang et al.,~\citeyear{2018PSMNet}), GwcNet (Guo et al.,~\citeyear{GWCNet}), RAFT (Lipson et al.,~\citeyear{lipson2021raft}) and ACVNet (Xu et al.,~\citeyear{2022ACVNet}), we rename the results after refining under our model as PSM-DV,  Gwc-DV, RAFT-DV and ACV-DV. We use completely the same training schedule as the corresponding work to make a fair comparison. The results show that we obtain 11\%, 5.2\%, 18\% and 4.2\% EPE improvement respectively while only adding 2\% parameters. The experiment shows that the redundant cost volume will interfere with the model training and our DiffuVolume can refine any of the existing volume-based methods by effectively filtering the volume. It is worth mentioning that adding our DiffuVolume into the PSMNet, it can directly outperform the CFNet (Shen et al.,~\citeyear{2021CFNet}) which is described in Table~\ref{tab:scene}, showing the great power of our filter mechanism. Overall, our DiffuVolume can be seen as a plug-and-play module that is uninterrupted by any methods but the essential flaw of the cost volume with few parameters addition.

\begin{table}[t]
\vspace{-8mm}
    \centering
    \caption{University of our DiffuVolume on Scene Flow dataset. DV means adding our DiffuVolume. Bold: Best.}
    \setlength{\tabcolsep}{0.8mm}
    \begin{tabular}{c|c|c|c|c}
        \hline
        Model & Year & EPE(px) $\downarrow$ & Bad1.0($\%$) $\downarrow$ & Params(M) \\
        \hline
        PSMNet & 2018 & 1.09 & 11.11 & 5.22 \\
        PSMNet-DV & - & \textbf{0.97} & \textbf{10.25} & 5.28 \\
        \hline
        GwcNet & 2019 & 0.77 & 8.03 & 6.91 \\
        GwcNet-DV & - & \textbf{0.73} & \textbf{7.89} & 6.97 \\
        \hline
        RAFT & 2021 & 0.72 & 7.30 & 11.12 \\
        RAFT-DV & - & \textbf{0.59} & \textbf{6.43} & 11.90 \\
        \hline
        ACVNet & 2022 & 0.48 & 5.04 & 7.17 \\
        ACVNet-DV  & - & \textbf{0.46} & \textbf{4.97} & 7.23 \\
        \hline
    \end{tabular}
    \centering
    \label{tab:plug}
\end{table}

\begin{table*}[ht]
    \centering
    \caption{The ablation study on the Scene Flow dataset. ``No iteration'' means the volume renewal and dense integration modules are not used, just using the standard Gaussian noise to filter the cost volume and predict the disparity map directly. The last two rows are ablated under 5 sampling steps. All the metrics are the lower the better. Bold: Best.} 
    \setlength{\tabcolsep}{2.5mm}
        \begin{tabular}{c|c|c|c|c|c}
        \hline
        Model & EPE(px) $\downarrow$ & D1($\%$) $\downarrow$ & \textgreater1px(\%) $\downarrow$ & \textgreater2px(\%) $\downarrow$ & \textgreater3px(\%) $\downarrow$ \\
        \hline
        No iteration & 0.54 & 1.82 & 6.04 & 3.26 & 2.35 \\
        1 sampling steps       & 0.48 & 1.68 & 5.15 & 2.88 & 2.12 \\
        2 sampling steps       & 0.47 & 1.68 & 5.13 & 2.88 & 2.12 \\
        4 sampling steps       & 0.47 & 1.66 & 5.07 & 2.85 & 2.10 \\
        5 sampling steps       & \textbf{0.46} & \textbf{1.62} & \textbf{4.97} & \textbf{2.77} & \textbf{2.04} \\
        10 sampling steps      & 0.47 & 1.62 & 5.00 & 2.79 & 2.05 \\
        \hline
        w/o volume renewal     & 0.47 & 1.66 & 5.04 & 2.83 & 2.09 \\
        w/o \blue{dense integration} & 0.48 & 1.62 & 5.29 & 2.86 & 2.07 \\
        \hline
        \end{tabular}
    \centering
    \label{tab:abalation}
\end{table*}

\begin{table*}[!th]
    \centering
    \caption{Result of different weight combinations, the weights means the result of step $1^{th}$, $2^{th}$, $3^{th}$, $4^{th}$, $5^{th}$ and previous work. We choose 0-0-0-0.2-0.3-0.5 as the final weights. Bold: Best.} 
    \begin{tabular}{c|c|c|c|c|c}
        \hline
        weight & 0-0-0-0-1-0 & 0.2-0.2-0.2-0.2-0.2-0 & 0.3-0.3-0.3-0-0-0.1 & 0-0-0-0-0.5-0.5 &
        0-0-0-0.2-0.3-0.5  \\
        EPE(px) & 0.49 & 0.48 & 0.47 & 0.47 & \textbf{0.46} \\
        \hline
    \end{tabular}
    \centering
    \label{tab:weight}
\end{table*}

\begin{table}[t]
    \centering
    \caption{The percentage of outliers at different steps, $\dagger$ means iterating with volume renewal.}
    \label{tab:renewal}
    \setlength{\tabcolsep}{1.5mm}
    \begin{tabular}{c|c|c|c|c|c}
        \hline
        step number & 1 & 2 & 4 & 5 & 10 \\
        \hline
        outliers(\%) & 49.24 & 37.09 & 30.74 & 29.86 & 29.63 \\
        outliers(\%) $\dagger$ & \textbf{45.52} & \textbf{32.14} & \textbf{24.32} & \textbf{23.25} & \textbf{23.06} \\
        \hline
        \end{tabular}
    \centering
\end{table}

\subsubsection{Ablation Study}
\label{sec:ablation}
We do all our ablation study on the Scene Flow dataset and use the architecture of ACVNet (Xu et al.,~\citeyear{2022ACVNet}) with two 3D stacked hourglass modules and 64-channel dimensions without further instructions. The results are all shown in Table~\ref{tab:abalation}. It is worth mentioning that due to the uniqueness of diffusion-based methods, all of the ablation modules are used in the inference stage.
\par

\noindent\textbf{Volume renewal.} As shown in Table~\ref{tab:renewal}, the number of outliers selected decreases with the number of iterations grows and the performance improves as shown in Table~\ref{tab:abalation}. The results show that without using the volume renewal, the number of outliers is greater, proving the effectiveness of the volume renewal module. It can also be seen as a rectified mechanism of our DiffuVolume. Outlier means that diffusion filter performs poorly at this point, a new initialization may rectify the filtering quality.

\par
\vspace{2mm}
\noindent\textbf{\blue{Dense Integration}.} \orange{The result in Table~\ref{tab:abalation} shows the effectiveness of dense integration module. We further} evaluate the different weight combinations of different sampling steps in Table~\ref{tab:weight}. Aggregating the results obviously improves the performance of the disparity map, and the later the steps, the more contribution to the final result. \red{However, in Fig.~\ref{fig:step}, although \blue{the} 5$^{th}$ step performs \blue{greatly} on the edge region, the prediction of the center region of the guitar is worse than the 1$^{st}$ step}. The results ensure that at each step, the diffusion filter may recover poorly on some points while greatly on other points. Through the weighted average, the overall trend will turn for the better. With the phenomenon in mind, we \textbf{reclaim} that the \blue{dense integration} module is essential for the diffusion model in \red{stereo matching}.
\par

\begin{table*}[t]
    \centering
    \caption{Comparing the performance of the image-denoising-based diffusion model and our DiffuVolume in stereo matching. The inference speed means the time for inferring one image. Bold: Best.}
    \resizebox{1.8\columnwidth}{!}{
    \begin{tabular}{c|c|c|c|c|c}
        \hline
        Model & Year & EPE(px) $\downarrow$ & Bad 1.0(\%) $\downarrow$ & Inference speed(s) $\downarrow$ & Params(M) $\downarrow$ \\
        \hline
        DDPM & 2020 & 0.59 & 6.06 & 265 & 60.07\\
        DDIM & 2020 & 0.63 & 6.13 & 1.21 & 60.07 \\
        DiffuVolume(ours) & - & \textbf{0.46} & \textbf{4.97} &  \textbf{1.11} & \textbf{7.23} \\
        \hline
    \end{tabular}
    }
    \centering
    \label{tab:diffusion}
\end{table*}

\begin{figure*}[t]
    \centering
    \includegraphics[width=0.95\linewidth]{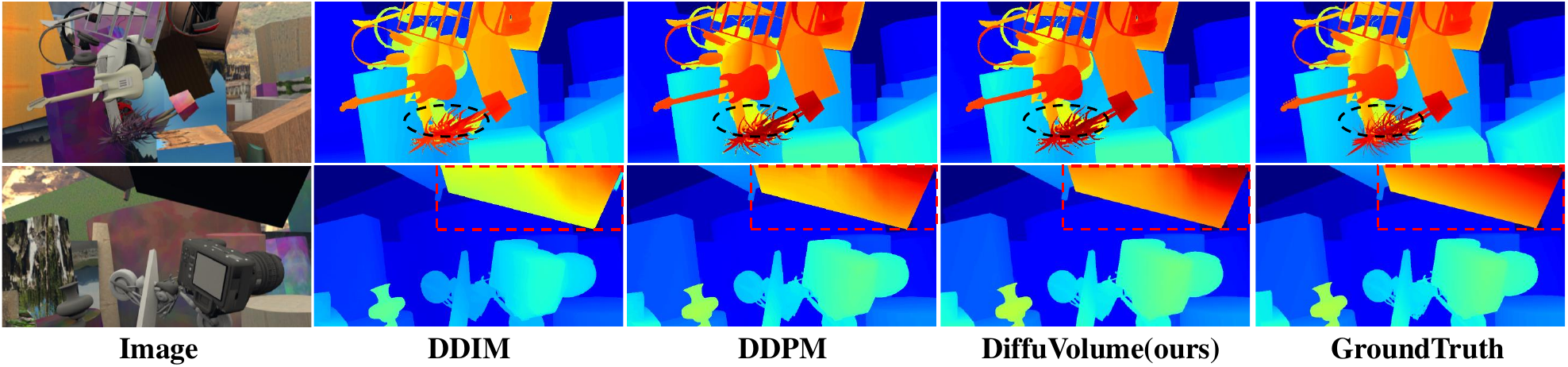}
    \caption{Visualization results of our DiffuVolume compared with traditional diffusion methods. The second and third columns are the results of DDIM and DDPM respectively, the fourth column is the result of our DiffuVolume and the last column is the ground truth map.}
    \label{fig:image_noise}
\end{figure*}
\vspace{2mm}
\noindent\textbf{Sampling step.} We evaluate the influence of different sampling steps on our model using all the proposed modules. The results represent that by only using one sampling step, our model can achieve comparable results with the ACVNet (Xu et al.,~\citeyear{2022ACVNet}) and as the sampling steps grow, the refinement performance goes better until five sampling steps, which not only ensures us that our DiffuVolume can filter some information in the cost volume which is redundant for model training but also represents that the diffusion filter is gradually approaching the original ground truth volume during the reverse process. This observation validates the reliability of our novel diffusion design. As for the performance drop at more steps, it is consistent with our observation in Sec~\ref{sec:4.3.1}, which ensures that diffusion model exists over-refining phenomenon and suitable sampling steps are significant.
\par

\begin{figure*}[t]
    \centering
    \includegraphics[width=0.95\linewidth]{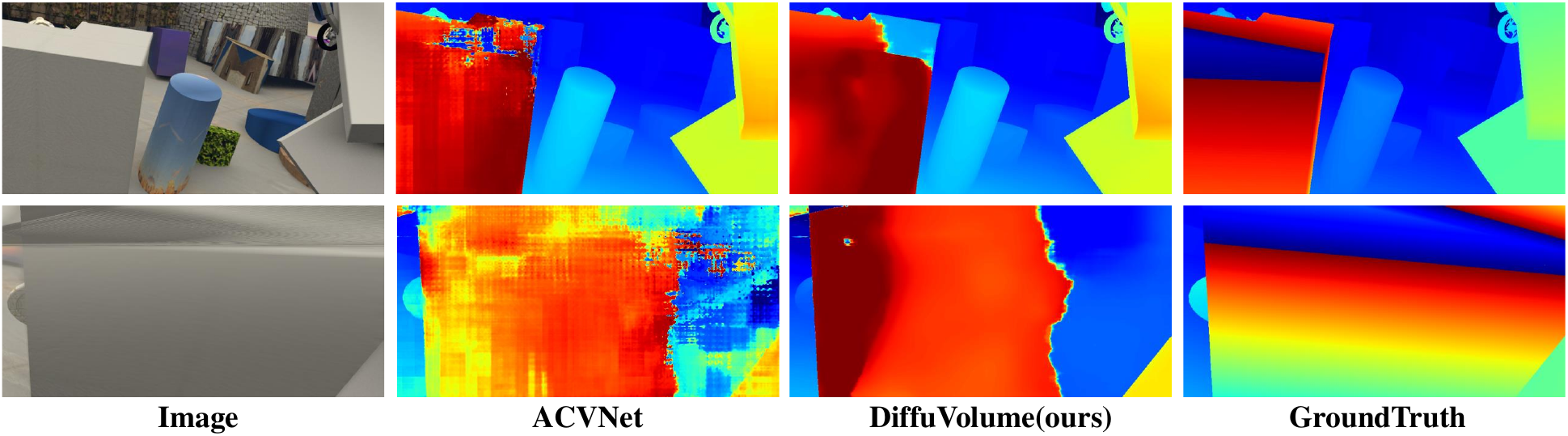}
    \caption{Visualization of the failing case on Scene Flow dataset.}
    \centering
    \vspace{-6mm}
    \label{fig:fail}
\end{figure*}

\begin{figure*}[t]
    \centering
    \includegraphics[width=0.95\linewidth]{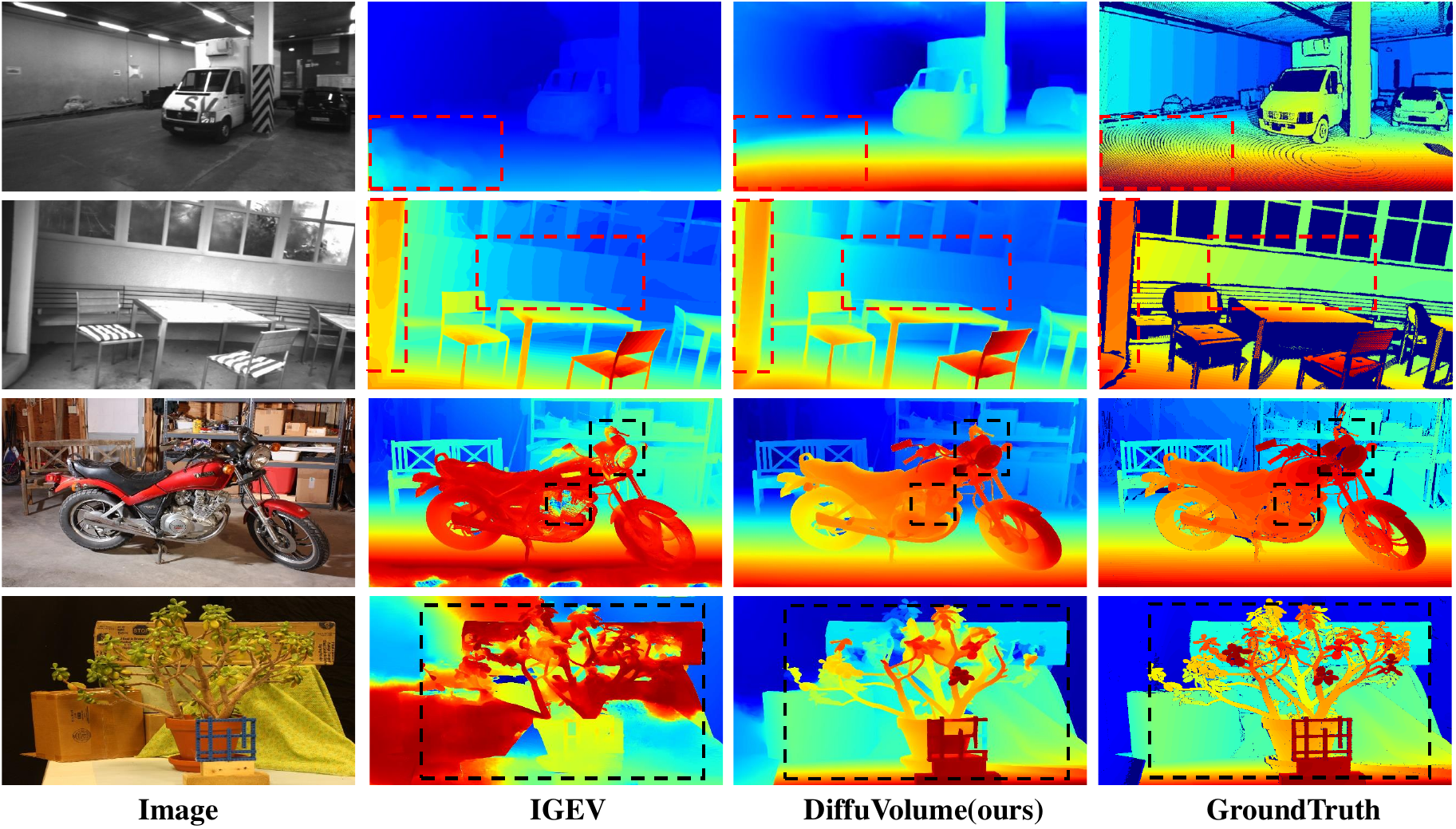}
    \caption{The visualization result of zero-shot generalization. The first two rows are tested on ETH3D and the rest on MIddlebury$\_$F. Our DiffuVolume performs better on edge regions and context information.}
    \label{fig:zero-shot}
\end{figure*}

\subsubsection{Failing Case}
\label{sec:fail}
In this section, we show the failing case in Fig.~\ref{fig:fail}. While the testing sample is extremely hard (\textit{e.g.}, the whole image is non-texture), neither ACVNet nor ours could predict the right disparity map. This is caused by extremely limited useful information in the cost volume. Although both ACVNet and our DiffVolume remove much redundant information in cost volume, \blue{both methods} ignore another view that how to use the useful information more effectively. This problem is essential in the non-texture image. Future work will focus on it.

\subsection{How could diffusion model be used reasonably in stereo matching}
To further prove the reliability of our novel application of the diffusion model, we conduct experiments to show that stereo matching is not suitable to be accomplished by trivially applying an existing image-denoising-based diffusion model. We implement the image-denoising-based diffusion stereo matching following DiffuStereo (Shao et al.,~\citeyear{2022diffustereo}). We add noise into the residual of the ground truth map and coarse disparity map obtained by ACVNet (Xu et al.,~\citeyear{2022ACVNet}), following the training schedule of DDPM (Ho et al.,~\citeyear{ddpm}). To constrain the model training, we add extensive conditions(\textit{e.g.}, edge map, original left image and coarse disparity map) to the U-Net. \orange{We further substitute the DDIM (Song et al.,~\citeyear{ddim}) sampling strategy for the DDPM sampling, which could accelerate the inference speed.}   
\par

The results are shown in Table~\ref{tab:diffusion}. We obtain 22\% EPE improvement compared with DDPM while accelerating the inference time by about 240 times. Compared with the DDIM sampling strategy, we achieved 27\% performance improvement while faster inference speed. We further visualize the disparity map compared with both methods as shown in Fig.~\ref{fig:image_noise}. The results of the DDIM method in the second row show that image-denoising-based diffusion stereo matching methods suffer from the \orange{disparity value} shift problem. It is caused by inaccurate pixel generation in the reverse process of diffusion. Even adding the condition to constrain the generation, this phenomenon still exists. Our DiffuVolume performs better on non-texture and edge regions which benefit from the information interaction in adjacent disparity levels.
\par
With the experiment results in mind, we argue that diffusion model has the potential for fusing into many dense prediction tasks. Firstly, the diffusion model can be designed to a task-specific module with slight parameters rise. Secondly, the iterative refinement idea in the reverse process is also a significant part of many computer vision tasks, which can be bridged to form a novel pipeline. Finally, the ability to recover the desired objective in a generative way may be a novel view for prediction-based tasks.
\par

\subsection{Zero-shot Generalization}
As the several real-world datasets are too small to train, the zero-shot generalization ability from large synthetic dataset to unseen real-world scenes is important. We embed our DiffuVolume into the RAFT-Stereo (Lipson et al.,~\citeyear{lipson2021raft}) and the result is shown in Table~\ref{tab:raft}. \blue{We compare with PSMNet (Chang et al.,~\citeyear{2018PSMNet}); GANet (Zhang et al.,~\citeyear{2019GA}); DSMNet (Zhang et al.,~\citeyear{dsmnet}); RAFT (Lipson et al.,~\citeyear{lipson2021raft}); CFNet (Shen et al.,~\citeyear{2021CFNet}) and IGEV (Xu et al.,~\citeyear{xu2023iterative})}. 
\par
\begin{table}[t]
    \centering
    \caption{Zero-shot generalization on real-world benchmarks. Errors are the percent of pixels with epe greater than the threshold, 3px for KITTI, 1px for ETH3D, 2px for Middlebury \textbf{in all regions}.}
    \setlength{\tabcolsep}{0.6mm}
    \begin{tabular}{cc|c|c|c|cc}
        \hline
        \multicolumn{2}{c|}{\multirow{2}{*}{Model}} & 
        \multirow{2}{*}{Year} & \multirow{2}{*}{KITTI15} & \multirow{2}{*}{ETH3D} & \multicolumn{2}{c}{Middlebury} \\
        \multicolumn{2}{c|}{} & {} & {} & {} & full & half  \\
        \hline
        \multicolumn{2}{c|}{PSMNet} & 2018 & 16.3 & 23.8 & 39.5 & 25.1  \\
        \multicolumn{2}{c|}{GANet} & 2019 & 11.7 & 14.1 & 32.2 & 20.3  \\
        \multicolumn{2}{c|}{DSMNet} & 2020 & 6.5 & 6.2 & 21.8 & 13.8  \\
        \multicolumn{2}{c|}{RAFT} & 2021 & 5.74 & 3.28 & 18.33 & 12.59  \\
        \multicolumn{2}{c|}{CFNet} & 2021 & 5.8 & 5.8 & 28.2 & - \\
        \multicolumn{2}{c|}{IGEV} & 2023 & 6.03 & 3.59 & 28.96 & \textbf{9.49} \\
        \multicolumn{2}{c|}{DiffuVolume(ours)} & - & \textbf{5.56} & \textbf{2.50} & \textbf{15.21} & 10.79 \\
        \hline
        \end{tabular}
    \label{tab:raft}
\end{table}

Our DiffuVolume improves the generalization ability of the model. In Fig.~\ref{fig:zero-shot}, we further compare the result of ours with IGEV-Stereo (Xu et al.,~\citeyear{xu2023iterative}), which encodes geometry and context information into all-pairs cost volume in RAFT-stereo. We perform better on the edge and detailed regions as the IGEV encodes the geometry information based on lightweight 4D cost volume, which exists redundant information. What's more, the max disparity range is set to 192 in the 4D cost volume of IGEV, when dealing with the high-resolution image (\textit{e.g.}, row 3$^{th}$, 4$^{th}$), the effect is diminished, while our diffusion filter still works. The experiments ensure that DiffuVolume can improve the generalization ability of volume-based models both in low-resolution datasets (\textit{i.e.}, ETH3D, KITTI) and high-resolution datasets (\textit{i.e.}, Middlebury).

\section{Conclusion}
\label{sec:conclu}
In this work, we have proposed DiffuVolume, a novel application for diffusion models in stereo matching as a cost volume filter. Our DiffuVolume eschews the traditional manner of directly adding noise into the image but \blue{integrates} the diffusion model into a task-specific module, which outperforms the traditional diffusion-based stereo matching method and accelerates the inference time by 240 times. What's more, our diffusion filter is a lightweight plug-and-play module that can embed into all of the existing volume-based methods (2\% parameters rise). We achieve state-of-the-art performance on five public benchmarks, \textit{i.e.}, Scene Flow, KITTI2012, KITTI2015, Middlebury and ETH3D compared with previous methods, especially rank 1$^{st}$ on the KITTI2012 leader board, 2$^{nd}$ on the KITTI2015 leader board \red{since} 15, July 2023.

\backmatter



\section*{Declarations}
All authors certify that they have no affiliations with or involvement in any organization or entity with any financial interest or non-financial interest in the subject matter or materials discussed in this manuscript.



\bibliography{sn-bibliography}

\end{document}